\documentclass[sigconf]{acmart}
\AtBeginDocument{%
  }

\setcopyright{acmlicensed}
\copyrightyear{202}
\acmYear{2018}
\acmDOI{XXXXXXX.XXXXXXX}
\acmConference[ACM MM]{Make sure to enter the correct
  conference title from your rights confirmation email}{2026}{Rio de Janeiro, Brazil}
\acmISBN{978-1-4503-XXXX-X/2018/06}

\usepackage{colortbl} % 行背景色支持
\usepackage{enumitem}

\usepackage{bm}            % 加粗数学符号
\usepackage{latexsym}      % LaTeX 符号支持

\usepackage{array}         % 高级表格列定义
\usepackage{pgf}           % 用于数学计算 (如 \gain 命令)

\usepackage[dvipsnames, table]{xcolor}

\usepackage{multirow}      % 表格跨行
\usepackage{makecell}      % 表格单元格换行
\usepackage{adjustbox}     % 调整盒子大小
\usepackage{tabularx}      % 自适应宽度表格
\usepackage{subcaption}    % 子图标题

\usepackage{algorithm}     % 算法浮动体
\usepackage{algorithmic}   % 算法伪代码
\usepackage{listings}      % 代码高亮

\usepackage{pifont}        % 符号 (如 \ding)
\usepackage{bbding}        % 符号
\usepackage{wasysym}       % 符号
\usepackage{utfsym}        % 符号
\usepackage{fontawesome}   % Font Awesome 图标
\usepackage{xspace}        % 智能空格
\usepackage{verbatim}      % 原样输出文本
\usepackage{soul}          % 文字高亮 (如 \hl)
\usepackage{tikz}          % 用于绘制图标

\definecolor{lightgrayblue}{HTML}{ECEFF1}  
\definecolor{lightblue}{HTML}{E3F2FD}      
\definecolor{lightpeach}{HTML}{FFF3E0}     
\definecolor{closedBg}{HTML}{F5F5F5}        
\definecolor{qwen15Bg}{HTML}{E8F4FC}        
\definecolor{qwen15Highlight}{HTML}{D4EBFA} 
\definecolor{qwen15Text}{HTML}{2E5F8C}      
\definecolor{qwen7Bg}{HTML}{FFF7FB}         
\definecolor{qwen7Highlight}{HTML}{FFEDF5}  
\definecolor{qwen7Text}{HTML}{B87090}       
\definecolor{keyColBg}{HTML}{FFF9E6}          

\definecolor{qwenGreenHighlight}{HTML}{E8F5E9}
\definecolor{qwenGreenBg}{HTML}{F1F8E9}         
\definecolor{qwenGreenText}{HTML}{2E7D32}       

\definecolor{kellygreen}{rgb}{0.3, 0.73, 0.09}
\definecolor{alizarin}{rgb}{0.82, 0.1, 0.26}
\definecolor{royalblue}{rgb}{0.25,0.41,1}

\definecolor{row-blue}{HTML}{F5FFFA}
\definecolor{row-green}{HTML}{F1F6EC}
\definecolor{row-yellow}{HTML}{FFFFF0}
\definecolor{row-pink}{HTML}{FFF5F5}
\definecolor{superlightgrey}{gray}{0.7}
\definecolor{row-best}{HTML}{F3E5F5}
\definecolor{row-second}{HTML}{E0F7FA}
\definecolor{row-blue1}{HTML}{E8F4FC}
\definecolor{row-pink1}{HTML}{FFF7FB}

\definecolor{best}{HTML}{FFE8D9}
\definecolor{second}{HTML}{DAE3F5}

\definecolor{narrative}{HTML}{E9C5C4}
\definecolor{event}{HTML}{FAF5BF}
\definecolor{atomic}{HTML}{CFF1B6}

\definecolor{forestgreen}{RGB}{34,139,34}
\definecolor{ashgrey}{RGB}{178,190,181}

\newcommand{\gain}[2]{%
    \pgfmathsetmacro{\delta}{(#1-#2)/#2*100}%
    \pgfmathprintnumber[fixed, precision=1, showpos]{\delta}\%%
}

\newcommand{\cmark}{{\color{kellygreen} \ding{51}}}
\newcommand{\xmark}{{\color{alizarin} \ding{55}}}

\newcommand{\gray}[1]{\textcolor{gray}{#1}}

\newcommand{\tie}[1]{\textcolor{gray}{\scriptsize (-)}}

\newcommand{\icoyes}{%
  \tikz[baseline=(char.base)]{
    \node[circle, fill=white, inner sep=0pt, outer sep=0pt] (char) {
      \textcolor{forestgreen}{\faCheckCircle}
    };
  }\xspace
}

\newcommand{\icono}{%
  \tikz[baseline=(char.base)]{
    \node[circle, fill=white, inner sep=0pt, outer sep=0pt] (char) {
      \textcolor{ashgrey}{\faTimesCircle}
    };
  }\xspace
}

\begin{document}

\title{WeaveEarth: Structured Evidence Construction and Reasoning for Training-Free UHR Remote Sensing Understanding}

\author{Xianzhi Ma}
\orcid{0009-0007-4909-0894}
\affiliation{%
  \institution{School of Frontier Sciences, Nanjing University}
  \city{Suzhou}
  \country{China}}
\email{xianzhima@smail.nju.edu.cn}

\author{Shujun Wang}
\orcid{0009-0002-6023-1874}
\affiliation{%
  \institution{School of Frontier Sciences, Nanjing University}
  \city{Suzhou}
  \country{China}}
\email{502025830005@smail.nju.edu.cn}

\author{Xiaohan Li}
\orcid{0009-0003-9131-7731}
\affiliation{%
  \institution{School of Frontier Sciences, Nanjing University}
  \city{Suzhou}
  \country{China}}
\email{602025830003@smail.nju.edu.cn}

\author{Hao Liu}    
\orcid{0000-0002-0411-5396}
\affiliation{%
  \institution{School of Frontier Sciences, Nanjing University}
  \city{Suzhou}
  \country{China}}
\email{haoliu@nju.edu.cn}

\author{Changhua Pei}  
\orcid{0000-0001-9288-4787}
\affiliation{%
  \institution{Computer Network Information Center, Chinese Academy of Sciences}
  \city{Beijing}
  \country{China}}
\email{chpei@cnic.cn}

\author{Jianhui Li}
\orcid{0009-0001-6253-9808}
\affiliation{%
  \institution{School of Frontier Sciences, Nanjing University}
  \city{Suzhou}
  \country{China}}
\email{lijh@nju.edu.cn}
\authornote{Corresponding author.}


\begin{abstract}
Ultra-High-Resolution (UHR) remote sensing image understanding requires Vision-Language Models (VLMs) to capture both the global scene layout and sparse yet task-critical local details under limited computational budgets. Existing methods mainly follow two paradigms. One is \emph{passive perception}, which relies on resolution expansion or token compression and may therefore discard fine-grained details. The other is \emph{active perception}, which depends on multi-round zooming and search, but suffers from high latency, contextual fragmentation, and error accumulation. We argue that a more effective path toward UHR understanding lies not in accessing more, but in organizing better. To this end, we propose WeaveEarth, a training-free framework that reformulates UHR understanding as a problem of structured evidence construction and reasoning under global context constraints. Specifically, WeaveEarth first employs \emph{Global-Aware Evidence Construction} to select a compact, low-redundancy, and spatially complementary \emph{Minimal Support Evidence Set}. It then introduces \emph{Structured Evidence Reasoning}, which weaves local evidence, spatial metadata, and relative topology into a unified reasoning interface, thereby enhancing the VLM's ability to perform global-local joint reasoning. Extensive experiments show that WeaveEarth consistently outperforms strong baselines and existing UHR methods across multiple UHR remote sensing benchmarks and multiple frozen VLM backbones. Code is available at \url{https://github.com/XianZhi-Ma/WeaveEarth}.
\end{abstract}



%%  \includegraphics[width=\textwidth]{sampleteaser}
%%  \caption{Seattle Mariners at Spring Training, 2010.}
%%  \Description{Enjoying the baseball game from the third-base
%%  seats. Ichiro Suzuki preparing to bat.}

\maketitle

\section{Introduction}

\begin{figure*}[t]
  \includegraphics[width=1\linewidth]{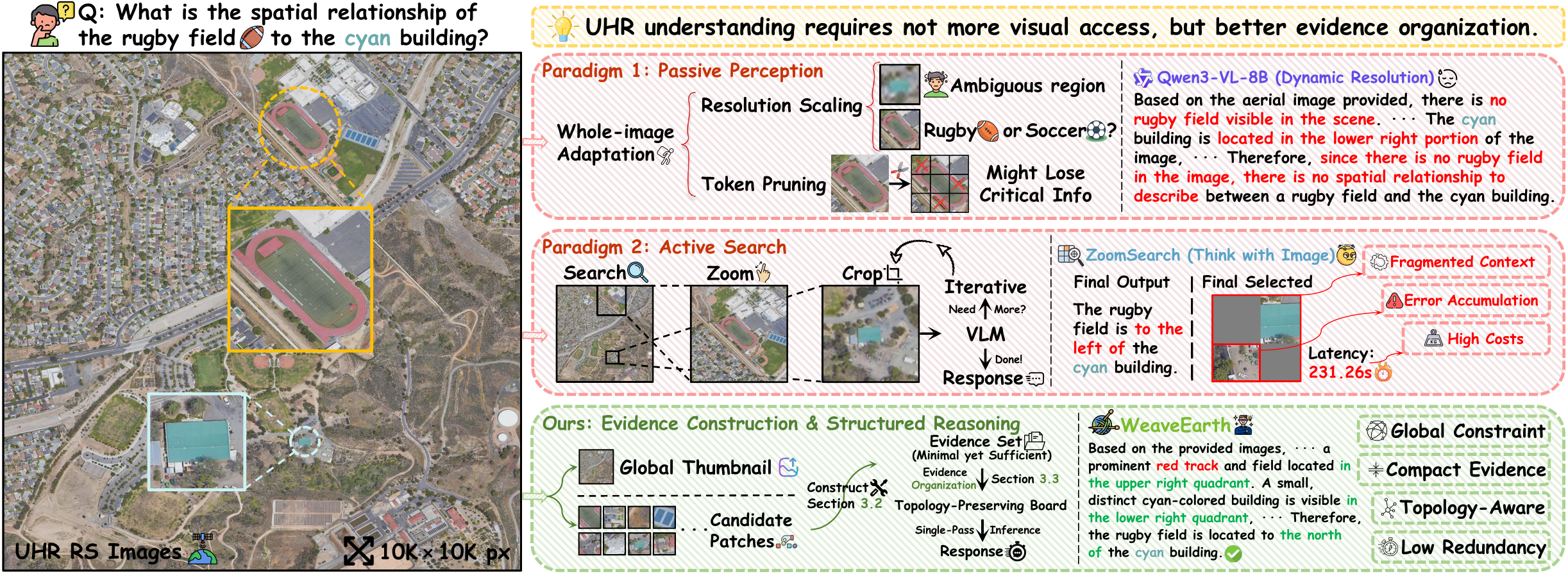}
  \caption{Paradigm comparison for UHR remote sensing understanding. Answering questions over UHR images requires jointly capturing the global layout and sparse task-critical local evidence. Existing methods mainly rely on \emph{passive whole-image adaptation}, which may lose critical details, or \emph{active multi-round search}, which often causes contextual fragmentation, error accumulation, and high latency. In contrast, WeaveEarth constructs a compact, globally constrained evidence set for structured reasoning, enabling effective UHR understanding through better evidence organization rather than expanded visual access.}
  \label{fig:1}
\end{figure*}

With the rapid advancement of Earth observation technologies, remote sensing (RS) image analysis is entering the era of Ultra-High-Resolution (UHR) imagery \citep{UHR1, UHR2, UHR3, UHR4, UHR5, UHR6}. Unlike ordinary natural images, UHR RS images exhibit a pronounced \emph{Macro-Micro Duality}: they span large-scale scene layouts while containing sparse, fine-grained, and task-critical local details. Enabling Vision-Language Models (VLMs) to understand such images is important for real-world applications such as environmental monitoring, fine-grained urban governance, and the monitoring and evaluation of the Sustainable Development Goals (SDGs) \citep{sdg}, where answering questions often depends on joint reasoning over global structure, local evidence, and cross-region relationships \citep{mm1, mm2, mm3, mm4, mm5, mm9}. Achieving such understanding under limited computational budgets is therefore not only important for practical deployment, but also poses a fundamental challenge for evidence-driven global-local reasoning in VLMs \citep{ImageRAG, geoeyes, TextBeforeVision}.

Existing UHR RS methods mainly follow two routes, yet both remain constrained by the trade-off between accuracy and efficiency. \emph{Passive perception} adapts the whole image through higher input resolution or token compression \citep{geollava, lrsvqa, ImageRAG}, but remains costly and may discard task-critical local evidence. \emph{Active perception} instead searches for evidence through iterative zooming and cropping \citep{zoomearth, zoomsearch, geoeyes, TextBeforeVision}, but often incurs high latency, contextual fragmentation, and additional search overhead, and may even require extra training for search or stopping policies. More importantly, sequential cropping can break long-range spatial dependencies, turning local evidence into isolated fragments that lack stable global grounding and are difficult to reason over jointly. As a result, a unified framework that can jointly support effective evidence selection, compact evidence organization, and reliable visual grounding under limited budgets is still lacking.

As illustrated in Figure~\ref{fig:1}, the bottleneck in UHR RS understanding lies not merely in expanding visual access, but in constructing and organizing the right evidence for reasoning. To address this gap, we propose \textbf{WeaveEarth}, a training-free framework for UHR RS understanding. Rather than feeding isolated local crops into the model, WeaveEarth weaves the global thumbnail, critical local evidence, and their spatial relationships into a unified structured evidence representation. Concretely, it first constructs a compact, low-redundancy, and spatially complementary \emph{Minimal Support Evidence Set} (MSES) under global context constraints, and then organizes it into a structured reasoning interface that preserves spatial grounding and relative topology. As a plug-and-play framework, WeaveEarth improves the efficiency and reliability of UHR reasoning without fine-tuning any backbone VLM parameters.

The main contributions of this work are summarized as follows:
\begin{enumerate}[leftmargin=*, itemsep=1pt, topsep=1pt]
    \item We propose \textbf{WeaveEarth}, a training-free framework that reformulates UHR RS understanding as structured evidence construction and joint reasoning under global context constraints.
    \item We introduce \textbf{Global-Aware Evidence Construction}, which builds a compact, low-redundancy, and spatially complementary \emph{Minimal Support Evidence Set} from a large candidate pool under limited budgets.
    \item We introduce \textbf{Structured Evidence Reasoning}, which uses \emph{Structured Evidence Metadata} and a \emph{Topology-Preserving Evidence Board} to preserve spatial grounding and relative topology for global-local reasoning.
    \item Extensive experiments on multiple UHR benchmarks and frozen VLM backbones demonstrate consistent gains, while ablations show that the improvements come from better evidence selection and organization rather than from expanded visual access.
\end{enumerate}
\section{Related Work}

\textbf{RS-VLMs.} In recent years, open-source VLMs have achieved remarkable progress in general visual understanding and multimodal reasoning, such as LLaVA v1.6 \citep{llava1.6} and IXC-2.5 \citep{ixc2.5}. Inspired by these advances, increasing efforts have sought to inject RS domain knowledge into general-purpose VLMs, extending remote sensing vision-language models (RS-VLMs) from image-level understanding to region-level and even pixel-level tasks \citep{RemoteCLIP, mm6, mm7, mm8, mm10, geomag, add9, add10, yl1, lk1, lk2, lk3}. Early methods such as RSGPT \citep{RSGPT} and RS-CapRet \citep{RS-CapRet} mainly focused on image-level captioning and scene understanding. Subsequent works, including GeoChat \citep{GeoChat}, EarthMarker \citep{EarthMarker}, SkyEyeGPT \citep{SkyEyeGPT}, EarthGPT \citep{EarthGPT}, and SkySense++ \citep{skysense++}, progressively extended RS-VLMs to region-level comprehension, localization, and ROI-aware understanding. More recently, GeoPix \citep{GeoPix}, GeoPixel \citep{GeoPixel}, and Falcon \citep{falcon} further pushed RS-VLMs toward pixel-level segmentation. Despite the increasingly fine-grained task granularity supported by RS-VLMs, most of these methods are still built upon a modeling paradigm with fixed input resolution and conventional visual token budgets.

\textbf{UHR RS Image Understanding.} Existing methods for UHR RS image understanding mainly follow two routes: \emph{passive perception} and \emph{active perception}. \emph{Passive perception} methods directly adapt to ultra-large images through higher input resolution, visual token compression, or retrieval-augmented mechanisms. Representative works include the text-guided dynamic pyramid token pruning strategy of Luo et al. \citep{lrsvqa}, as well as ImageRAG \citep{ImageRAG} and GeoLLaVA-8K \citep{geollava}. Although these methods usually maintain a relatively simple inference pipeline without explicit multi-round interaction, they still rely on costly large-input processing and often sacrifice fine-grained local evidence during compression or retrieval. In contrast, \emph{active perception} methods formulate UHR understanding as searching for evidence before reasoning, dynamically selecting task-relevant regions through zooming, searching, or agentic interaction. Representative works include ZoomSearch \citep{zoomsearch}, ZoomEarth \citep{zoomearth}, the stage-wise knowledge-injected RL strategy \citep{add1, add2, add3, add4, add5, add6, add7, add8, yl2, yl3} of Wang et al. \citep{TextBeforeVision}, and GeoEyes \citep{geoeyes}. While more effective at locating relevant regions, they typically incur repeated interaction, higher latency, and fragmented spatial context.

\section{WeaveEarth}

\begin{figure*}[t]
  \includegraphics[width=1\linewidth]{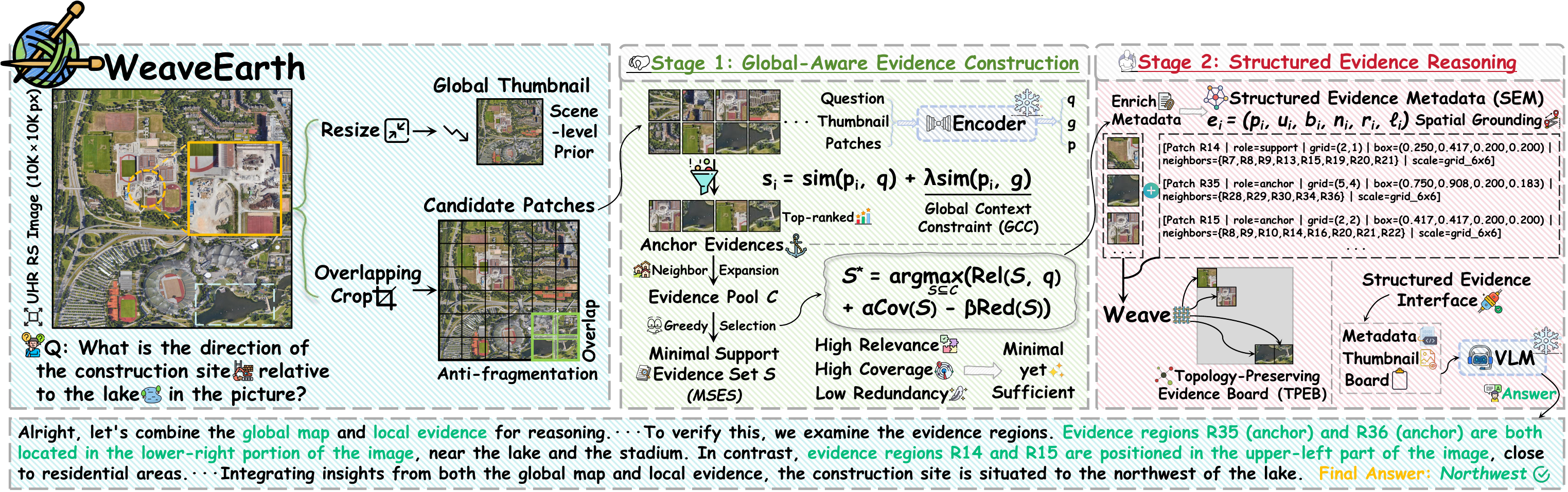}
  \caption{Overview of WeaveEarth. Stage~1 (\emph{Global-Aware Evidence Construction}) uses a global thumbnail and overlapping local crops to construct a compact, globally constrained \emph{Minimal Support Evidence Set}. Stage~2 (\emph{Structured Evidence Reasoning}) enriches selected patches with \emph{Structured Evidence Metadata} and organizes them into a \emph{Topology-Preserving Evidence Board} to preserve spatial grounding and relative topology. Together with the global thumbnail, they form a unified structured evidence interface for a frozen VLM, enabling global-local joint reasoning under limited visual budgets.}
  \label{fig:2}
\end{figure*}

\subsection{Framework Overview}
UHR RS understanding requires the model to jointly capture large-scale global layout and sparse yet answer-critical local evidence \citep{geollava, ImageRAG}. To address this challenge, we propose \textbf{WeaveEarth}, a structured evidence construction and reasoning framework for UHR RS understanding. Rather than treating UHR understanding as a problem of stronger whole-image perception, WeaveEarth reformulates it as structured evidence construction and joint reasoning under global context constraints. Instead of modifying the internal architecture of a frozen VLM, WeaveEarth redesigns its inference input interface through structured evidence composition.

As shown in Figure~\ref{fig:2}, WeaveEarth consists of two tightly coupled stages, corresponding to two core questions: how to construct critical evidence and how to organize evidence for reasoning. \textbf{Global-Aware Evidence Construction} builds a compact, low-redundancy, and spatially complementary \emph{Minimal Support Evidence Set} under global context constraints. \textbf{Structured Evidence Reasoning} then organizes the selected evidence together with spatial metadata and the global thumbnail into a unified structured evidence interface, enabling a frozen VLM to perform stable joint reasoning between global scene priors and task-critical local evidence. We next describe these two components in detail.

\subsection{Global-Aware Evidence Construction}
Instead of performing high-resolution reasoning over the entire image, we first divide the original image into a set of overlapping local regions to form a candidate evidence pool. This design is intended to avoid the case where a critical target lies on a region boundary, which would make it incomplete under non-overlapping partitioning. Meanwhile, we construct a low-resolution global thumbnail from the image to provide a scene-level contextual prior at low cost. The motivation is that, in UHR RS images, critical evidence is often distributed over only a few local regions, while the semantic value of these regions depends on the global scene layout in which they are situated. Therefore, the selection of local evidence should not rely solely on the question text, but should also be jointly constrained by the global context.

Formally, let the original image be divided into $N$ local regions $\{p_i\}_{i=1}^{N}$, with the question denoted as $q$ and the global thumbnail as $g$. We first employ a cross-modal semantic encoder to extract semantic representations for the local regions, the question text, and the global thumbnail, respectively. Then, for each candidate region, we compute a joint relevance score $s_i$ with respect to both the question and the global thumbnail:
\begin{equation}
s_i = \operatorname{sim}(p_i, q) + \lambda \operatorname{sim}(p_i, g),
\end{equation}
where $\operatorname{sim}(\cdot,\cdot)$ denotes the similarity between normalized features, and $\lambda$ balances the contributions of question semantics and the global scene prior. Unlike local patch retrieval based solely on the question, this design explicitly introduces a \emph{Global Context Constraint} (GCC), requiring candidate regions to be not only semantically relevant to the question, but also consistent with the scene layout of the entire image. Our key insight is that, in UHR RS images, the importance of a local region depends not only on whether it resembles the target described in the question, but also on whether it lies within a reasonable global context.

We first select the top-scoring candidate regions as anchor evidence. To alleviate the boundary fragmentation caused by local cropping, we further perform local neighborhood expansion around the anchor evidence, incorporating spatially adjacent regions into the candidate evidence pool. This step is not intended merely to increase the number of evidence regions, but rather to preserve necessary contextual information, allowing the model to perceive the spatial dependencies between local targets and their surrounding regions during subsequent reasoning.

However, relying solely on top-$k$ retrieval is still insufficient for high-quality reasoning. Directly selecting the highest-scoring regions often leads to two problems. First, the candidate regions may be highly redundant, with multiple patches repeatedly covering the same location or object. Second, regions that are semantically relevant are not necessarily spatially complementary, meaning that the resulting evidence set may be relevant but not necessarily sufficient. To address this issue, we further construct a \emph{Minimal Support Evidence Set} (MSES) from the candidate evidence pool, with the goal of preserving a small number of key regions that are sufficient to support answer prediction.

Formally, given a candidate region set $\mathcal{C}$, we aim to select a compact support set $\mathcal{S} \subseteq \mathcal{C}$ that simultaneously satisfies three requirements:
\begin{enumerate}[leftmargin=*, itemsep=1pt, topsep=2pt]
    \item high semantic relevance to the question;
    \item strong spatial complementarity, so as to cover richer regional information;
    \item low visual redundancy, avoiding repeated transmission of the same evidence across multiple regions.
\end{enumerate}

Accordingly, we formulate support set construction as the following objective:
\begin{equation}
\mathcal{S}^{*}=\arg\max_{\mathcal{S}\subseteq\mathcal{C}}
\mathrm{Rel}(\mathcal{S},q)+\alpha \mathrm{Cov}(\mathcal{S})-\beta \mathrm{Red}(\mathcal{S}),
\end{equation}
where $\mathcal{S}^*$ denotes the optimal support set, $\mathrm{Rel}$ denotes the overall relevance between the support set and the question, $\mathrm{Cov}$ measures spatial coverage and complementarity, and $\mathrm{Red}$ quantifies the redundancy within the evidence set. In practice, we adopt a training-free greedy strategy to progressively select candidate regions by jointly considering semantic scores, spatial coverage gains, and overlap penalties, thereby constructing a compact, low-redundancy, and spatially complementary support set. Since this process only involves lightweight ranking and set updating over a small number of candidate regions, its additional computational overhead is relatively small compared with the final VLM inference.

By unifying candidate retrieval and support set compression into a single evidence construction process under global context constraints, WeaveEarth substantially reduces the input scale for subsequent reasoning while preserving evidence quality, thereby laying the foundation for structured evidence organization and final joint reasoning.

\subsection{Structured Evidence Reasoning}
After the \textit{Global-Aware Evidence Construction} stage, WeaveEarth has already constructed a compact, low-redundancy, and spatially complementary \textit{MSES} from a large pool of local regions. However, merely obtaining a set of relevant patches is still insufficient for stable UHR reasoning. The reason is that, if these local regions are directly fed into the model as independent visual fragments, the model can often perceive only their local appearance, while failing to understand their spatial roles within the full image and the relative relationships among different regions. For example, for position-sensitive questions such as ``which object is on the left/right'' or ``which target is closer to the center,'' the model cannot make accurate judgments when only local appearance is available without global positional context. In UHR RS images, the answer to many questions depends not only on whether a local target exists, but also on where it is located, which regions it is adjacent to, and what semantic role it plays in the overall layout. Therefore, the second stage of WeaveEarth is not merely to \emph{use} the evidence, but to further organize it into a structured evidence representation that is suitable for global-local joint reasoning in VLMs.

Specifically, for each region in the MSES, we preserve not only its local visual content, but also explicitly encode its spatial attributes in the original image, including its normalized bounding box, relative grid position, adjacency relationships, and role information within the evidence set. In this way, each local region is no longer treated as a simple cropped image, but is instead represented as a structured evidence unit with explicit spatial semantics. We refer to the associated structured metadata as \textit{Structured Evidence Metadata} (SEM). Formally, given the $i$-th region in the support set, we represent it as
\begin{equation}
e_i = (p_i, u_i, b_i, n_i, r_i, \ell_i),
\end{equation}
where $e_i$ denotes the structured representation of the $i$-th evidence unit, $p_i$ denotes the local image content, $u_i$ denotes the region identifier, $b_i$ denotes its normalized spatial location in the full image, $n_i$ denotes its adjacency relationships with other evidence regions, $r_i$ denotes its role attribute within the current evidence set, and $\ell_i$ denotes its grid/scale information. As illustrated in Stage 2 of Figure~\ref{fig:2}, SEM provides a concrete structured representation for each evidence unit. Such a representation naturally endows each local region with both positional and relational information, preventing it from degenerating into an isolated visual fragment detached from context when fed into the model.

\begin{table*}[t]
  \centering
    \tiny
  \resizebox{\textwidth}{!}{
   \renewcommand{\arraystretch}{0.72}
    \begin{tabular}{lcccccccccc}
     \toprule
% 第一行
\multirow{2}{*}[-0.8ex]{\makecell{\bfseries Method}} & 
\multirow{2}{*}[-0.8ex]{\makecell{\bfseries \#Params}} & 
\multirow{2}{*}[-0.8ex]{\makecell{\bfseries Training \\ \bfseries -free}} &
\multicolumn{4}{c}{\bfseries LRS-VQA} & 
\multicolumn{4}{c}{\bfseries MME-RealWorld} \\
% 画线技巧
\cmidrule(lr){4-7} \cmidrule(lr){8-11} 
% 第二行
& & & \bfseries FAIR & \bfseries Bridge & \bfseries STAR & \bfseries Avg. & \bfseries Color & \bfseries Count & \bfseries Position & \bfseries Avg.  \\ 
\midrule
% --- Commercial VLMs ---
\multicolumn{11}{c}{\cellcolor{row-green}\gray{\textit{\textbf{Commercial VLMs}}}}\\
GPT-4o-mini \citep{gpt4omini} & - & \icono & 18.67 & 31.99 & 25.85 & 24.51 & 5.82 & 2.61 & 11.54 & 6.69 \\
GPT-4o \citep{gpt4o} & - & \icono & 22.15 & 31.84 & 27.40 & 26.42 & 34.18 & 15.17 & 37.07 & 28.81 \\
Claude-3.7-Sonnet \citep{claude3.7} & - & \icono & 14.28 & 28.52 & 15.03 & 16.75 & 32.90 & 19.42 & 28.57 & 27.02 \\
\midrule
% --- Open-Source General VLMs ---
\rowcolor{row-yellow}\multicolumn{11}{c}{\gray{\textit{\textbf{Open-Source General VLMs}}}}\\
Qwen2.5-VL \citep{qwen2.5vl} & 7B & \icono & 22.85 & 36.94 & 25.78 & 26.49 & 45.94 & 21.05 & 55.23 & 40.90 \\
% Qwen3-VL: 次好 (下划线) -> row-second
Qwen3-VL \citep{qwen3vl} & 8B & \icono & 24.61 & \underline{38.15} & \underline{27.53} & 28.16 & 48.07 & 21.35 & 56.41 & 42.11 \\
LLaVA-OV \citep{llavaov} & 7B & \icono & 20.61 & 35.11 & 26.08 & 25.69 & 26.81 & 26.14 & 27.57 & 26.85 \\
% IXC-2.5: 次好 (下划线) 和 最好 (加粗)
IXC-2.5 \citep{ixc2.5} & 7B & \icono & \underline{25.25} & \textbf{38.41} & 27.30 & \cellcolor{row-second}\underline{28.27} & 45.34 & 17.62 & 44.95 & 36.12 \\
LLaVA-v1.5 \citep{llava1.5} & 7B & \icono & 18.76 & 30.70 & 22.63 & 22.60 & 23.11 & 16.88 & 26.25 & 22.12 \\
LLaVA-v1.6 \citep{llava1.6} & 7B & \icono & 19.55 & 31.97 & 23.58 & 23.55 & 24.51 & 18.03 & 27.49 & 23.39 \\
CogVLM2 \citep{CogVLM2} & 8B & \icono & - & - & - & - & 37.69 & 18.35 & 29.99 & 28.76 \\
ZoomEye \citep{ZoomEye} & 7B & \icono & 22.53 & 34.48 & 26.27 & 26.30 & 48.54 & 17.76 & 31.18 & 32.61 \\
\midrule
% --- RS-VLMs ---
\rowcolor{row-pink}\multicolumn{11}{c}{\gray{\textit{\textbf{RS-VLMs}}}} \\
RSUniVLM \citep{RSUniVLM} & 0.5B & \icono & 21.02 & 32.61 & 24.72 & 26.07 & 28.97 & 19.00 & 28.33 & 25.48 \\
GeoChat \citep{GeoChat} & 7B & \icono & 20.18 & 24.54 & 13.75 & 17.30 & 23.11 & 15.66 & 25.06 & 21.32 \\
VHM \citep{VHM} & 7B & \icono & 22.59 & 34.40 & 26.75 & 26.57 & 20.32 & 16.80 & 35.24 & 24.18 \\
\midrule
% --- UHR RS Understanding Methods ---
\rowcolor{row-blue}\multicolumn{11}{c}{\gray{\textit{\textbf{UHR RS Understanding Methods}}}} \\
Luo et al. (2025) \citep{lrsvqa} & 7B & \icono & 22.97 & 36.89 & 27.48 & 27.45 & - & - & - & 39.04 \\
% ZoomEarth: 最好 (加粗)
ZoomEarth \citep{zoomearth} & 3B & \icono & - & - & - & - & \underline{49.69} & \textbf{31.66} & 50.14 & \cellcolor{row-second}\underline{43.93} \\
GeoLLaVA-8K \citep{geollava} & 7B & \icono & 20.81 & 32.73 & 24.56 & 24.58 & 27.92 & 22.27 & 34.90 & 28.41 \\
% ZoomSearch: 次好 (下划线)
ZoomSearch \citep{zoomsearch} & 7B & \icoyes & 21.83 & 32.33 & 26.05 & 25.65 & 48.62 & 20.05 & \underline{59.83} & 43.02 \\
% WeaveEarth (Ours): 混合了最好 (加粗) 和 次好 (下划线)
\cellcolor{row-best}\textbf{WeaveEarth (Ours)} & 8B & \icoyes & \textbf{30.99} & 26.08 & \textbf{36.68} & \cellcolor{row-best}\textbf{33.38} & \textbf{51.39} & \underline{27.65} & \textbf{62.60} & \cellcolor{row-best}\textbf{47.38} \\

      \bottomrule
    \end{tabular}%
}
  \caption{Main results on two UHR RS benchmarks, LRS-VQA and MME-RealWorld. The reported Avg. is the sample-size-weighted average accuracy (\%) across subcategories, ensuring a fairer overall comparison under category imbalance. Rankings are computed column-wise, with the best result highlighted in purple and bold, and the second-best result highlighted in blue and underlined. The symbol ``-'' indicates that the corresponding category result was not reported in the original paper.}
  \label{table:1}
\end{table*}

Based on this, we further construct a \textit{Topology-Preserving Evidence Board} (TPEB) to organize the discrete evidence regions into a unified visual input. Unlike simply arranging patches according to retrieval order or tiling them on a fixed grid, WeaveEarth preserves, as much as possible under a compact visual layout, the coarse-grained relative topology and adjacency relationships of the support set in the original image, so that spatially adjacent or structurally related local regions remain close to each other on the evidence board. The key motivation is that, for spatial relations, distribution patterns, and multi-region joint reasoning in RS scenes, textual prompts alone are often insufficiently stable for conveying positional relationships, whereas preserving relative topology and adjacency consistency provides VLMs with a more reliable spatial inductive bias. In other words, WeaveEarth not only \emph{selects} evidence, but also \emph{weaves} local evidence into a visually structured reasoning interface through topology-preserving evidence organization.

It is worth noting that SEM and TPEB play complementary but distinct roles in WeaveEarth. SEM addresses the question of what each evidence region is, where it is located, and how it relates to other regions, by providing each patch with explicit and readable spatial grounding. In contrast, TPEB addresses how multiple evidence regions should be jointly perceived by the model as a whole, by preserving relative adjacency and layout consistency to provide a more stable spatial organization for VLMs. Together, they form the structured evidence reasoning interface for a frozen VLM.

Finally, WeaveEarth feeds three types of information into the frozen VLM:
\begin{enumerate}[leftmargin=*, itemsep=1pt, topsep=2pt]
    \item the global thumbnail that provides a scene-level prior;
    \item the topology-preserving evidence board constructed from the MSES;
    \item the structured evidence metadata corresponding to each evidence unit.
\end{enumerate}

Under this input interface, the model no longer needs to search for key regions by itself within the full UHR image, but can instead directly perform joint reasoning over a limited amount of critical local evidence under global context constraints. The core insight of this design is that, for UHR RS image understanding, the key challenge is not merely to \emph{find} evidence, but to organize evidence into a unified representation that preserves spatial structure and is amenable to model reasoning.

\section{Experiments}
\subsection{Experimental Setup}
\textbf{Benchmarks and Metrics.} We evaluate WeaveEarth on three representative UHR RS image understanding benchmarks, including one visual question answering dataset, LRS-VQA (7,333 questions) \citep{lrsvqa}, and two multiple-choice benchmarks, MME-RealWorld (3,738 questions) \citep{mme} and XLRS-Bench (3,080 questions) \citep{XLRS}. To ensure fair evaluation, we follow the official data splits and evaluation protocols of each benchmark, and uniformly adopt accuracy as the evaluation metric. For benchmarks with multiple subcategories, the overall Avg. is computed as the sample-size-weighted average accuracy across all subcategories, thereby ensuring a fairer overall comparison under category imbalance.

\begin{table*}[t]
  \centering
      \tiny
  \resizebox{\textwidth}{!}{
   \renewcommand{\arraystretch}{0.72}
    \begin{tabular}{lccccccccccc}
\toprule
% 第一行
\multirow{2}{*}[-0.8ex]{\makecell{\bfseries Method}} & 
\multirow{2}{*}[-0.8ex]{\makecell{\bfseries \#Params}} & 
\multirow{2}{*}[-0.8ex]{\makecell{\bfseries Training \\ \bfseries -free}} &
\multicolumn{4}{c}{\bfseries Perception} & 
\multicolumn{4}{c}{\bfseries Reasoning} & 
\multirow{2}{*}[-0.8ex]{\bfseries Avg.} \\ 
% 画线技巧
\cmidrule(lr){4-7} \cmidrule(lr){8-11} 
% 第二行
& & & \bfseries Cnt & \bfseries SC & \bfseries OSR & \bfseries OP & \bfseries Plan & \bfseries AR & \bfseries CR & \bfseries SR &  \\ 
\midrule
% --- Commercial VLMs ---
\multicolumn{12}{c}{\cellcolor{row-green}\gray{\textit{\textbf{Commercial VLMs}}}}\\
GPT-4o-mini \citep{gpt4omini} & - & \icono & 24.38 & 39.33 & 23.60 & 45.60 & 29.00 & 71.00 & 50.50 & 6.67 & 40.16 \\
GPT-4o \citep{gpt4o} & - & \icono & 28.75 & 40.33 & 24.60 & 10.78 & 35.00 & 73.00 & 49.00 & 20.00 & 22.30 \\
Claude-3.7-Sonnet \citep{claude3.7} & - & \icono & 25.00 & 43.00 & 27.60 & 41.33 & 35.00 & 65.00 & 55.50 & 28.33 & 39.65 \\
\midrule
% --- Open-Source General VLMs ---
\rowcolor{row-yellow}\multicolumn{12}{c}{\gray{\textit{\textbf{Open-Source General VLMs}}}}\\
Qwen2.5-VL \citep{qwen2.5vl} & 7B & \icono & 30.63 & 40.33 & 30.80 & 43.31 & 32.00 & 67.00 & 62.50 & 43.33 & 41.98 \\
% Qwen3-VL: 次好
Qwen3-VL \citep{qwen3vl} & 8B & \icono & 32.50 & 41.67 & 31.40 & 43.92 & 34.00 & 70.00 & 63.50 & \underline{46.33} & 42.92 \\
% LLaVA-OV: 次好
LLaVA-OV \citep{llavaov} & 7B & \icono & 31.25 & 35.67 & 25.20 & 45.42 & 24.00 & \underline{76.00} & 59.50 & 43.33 & 41.62 \\
% IXC-2.5: 次好
IXC-2.5 \citep{ixc2.5} & 7B & \icono & 31.88 & 37.67 & 26.00 & 22.11 & \underline{41.00} & 72.00 & 64.00 & 36.67 & 30.00 \\
LLaVA-v1.5 \citep{llava1.5} & 7B & \icono & 22.50 & 19.67 & 23.00 & 20.18 & 39.00 & 35.00 & 34.00 & 30.00 & 22.89 \\
LLaVA-v1.6 \citep{llava1.6} & 7B & \icono & 24.38 & 21.33 & 23.80 & 21.27 & 40.00 & 35.00 & 34.50 & 33.33 & 24.00 \\
% CogVLM2: 次好
CogVLM2 \citep{CogVLM2} & 8B & \icono & \underline{36.88} & 46.33 & \underline{36.20} & 36.87 & 34.00 & 69.00 & 60.50 & - & 40.23 \\
LLaVA-Next \citep{llavanext} & 8B & \icono & 33.13 & 36.00 & 30.00 & 42.71 & 27.00 & 69.00 & 57.00 & 35.00 & 40.62 \\
% InternVL3: 次好 和 最好
InternVL3 \citep{InternVL3} & 8B & \icono & \underline{36.88} & 40.67 & 25.20 & \underline{46.80} & 36.00 & \textbf{77.00} & \underline{66.00} & 21.67 & \cellcolor{row-second}\underline{43.57} \\
\midrule
% --- RS-VLMs ---
\rowcolor{row-pink}\multicolumn{12}{c}{\gray{\textit{\textbf{RS-VLMs}}}} \\
GeoChat \citep{GeoChat} & 7B & \icono & 22.50 & 12.67 & 24.20 & 24.28 & 10.00 & 33.00 & 32.00 & - & 23.35 \\
\midrule
% --- UHR RS Understanding Methods ---
\rowcolor{row-blue}\multicolumn{12}{c}{\gray{\textit{\textbf{UHR RS Understanding Methods}}}} \\
ZoomEarth \citep{zoomearth} & 3B & \icono & - & - & - & - & - & - & - & - & 40.97 \\
% GeoLLaVA-8K: 混合 最好 和 次好
GeoLLaVA-8K \citep{geollava} & 7B & \icono & 35.00 & \textbf{62.00} & 35.80 & 37.71 & \textbf{65.00} & 68.00 & 63.50 & \textbf{51.67} & 43.44 \\
% WeaveEarth (Ours): 混合 最好 和 次好
\cellcolor{row-best}\textbf{WeaveEarth (Ours)} & 8B & \icoyes & \textbf{37.50} & \underline{50.33} & \textbf{36.40} & \textbf{47.41} & 36.00 & 73.00 & \textbf{67.50} & \underline{46.33} & \cellcolor{row-best}\textbf{47.14} \\
\bottomrule
    \end{tabular}%
  }
  % 修改 caption 为适合双栏的长文本排版
\caption{Results on the UHR RS benchmark XLRS-Bench. The reported Avg. is the sample-size-weighted average accuracy (\%) across subcategories, ensuring a fairer overall comparison under category imbalance. The subcategories include Counting (Cnt), Scene Classification (SC), Object Spatial Relationship (OSR), Object Properties (OP), Planning (Plan), Anomaly Reasoning (AR), Complex Reasoning (CR), and Spatiotemporal Reasoning (SR). Rankings are computed column-wise, with the best result highlighted in purple and bold, and the second-best result highlighted in blue and underlined.}
  \label{table:2}
\end{table*}

\noindent\textbf{Implementation Details.} All experiments are conducted on an NVIDIA A100 80GB GPU, with CUDA 12.4, Python 3.11.11, and PyTorch 2.5.1. To validate the plug-and-play capability of our model, in the main experiments we instantiate it on three representative open-source VLM backbones, Qwen3-VL-8B \citep{qwen3vl}, LLaVA-v1.6-7B \citep{llava1.6}, and IXC-2.5-7B \citep{ixc2.5}. All backbones are kept frozen, without any parameter fine-tuning. Unless otherwise specified, all remaining ablation studies, efficiency analyses, and case studies are conducted using Qwen3-VL-8B. More details about the experimental environment and hyperparameter settings are provided in the appendix.

\subsection{Main Results}

As shown in Table~\ref{table:1}, we evaluate WeaveEarth on two UHR remote sensing benchmarks, LRS-VQA and MME-RealWorld, where it achieves the best overall performance on both datasets, with average accuracies of 33.38\% and 47.38\%, respectively. Specifically, WeaveEarth surpasses the previous best UHR image understanding methods, outperforming Luo et al. (2025) \citep{lrsvqa} on LRS-VQA by 5.93 percentage points (from 27.45\% to 33.38\%) and ZoomEarth \citep{zoomearth} on MME-RealWorld by 3.45 percentage points (from 43.93\% to 47.38\%), establishing new state-of-the-art results.

Compared with the corresponding baselines, WeaveEarth consistently improves performance across multiple subcategories, indicating that its advantage is not limited to a single question type. These results suggest that, compared with brute-force whole-image perception or multi-round search, explicitly constructing and organizing the key evidence that supports the answer is more effective.

We further report the fine-grained results on XLRS-Bench in Table~\ref{table:2}. WeaveEarth achieves the best weighted average accuracy of 47.14\%, and attains the best performance on key sub-tasks such as \emph{Object Spatial Relationship} and \emph{Complex Reasoning}, which rely more heavily on spatial structure and reasoning ability. This observation is consistent with our design motivation: the strength of WeaveEarth lies not in exposing the model to more visual content, but in providing a compact and spatially organized evidence interface that better supports fine-grained recognition and multi-region reasoning. In this sense, the results on XLRS-Bench lend empirical support to our method design, showing that the \emph{Minimal Support Evidence Set} and \emph{Topology-Preserving Evidence Board} indeed help improve the key capabilities required by UHR benchmarks.

\begin{table*}[t]
\tiny
  \centering
  \resizebox{\textwidth}{!}{
   \renewcommand{\arraystretch}{0.93}
    \begin{tabular}{lcccccccccc}
\toprule
% 第一行表头
\multirow{2}[2]{*}{\bfseries Variant} & 
\multirow{2}[2]{*}{\bfseries\makecell{Con\\straint}} & 
\multirow{2}[2]{*}{\bfseries\makecell{Support \\ Set}} & 
\multirow{2}[2]{*}{\bfseries\makecell{Meta \\ data}} & 
\multirow{2}[2]{*}{\bfseries\makecell{Evidence \\ Board}} & 
\multirow{2}[2]{*}{\bfseries\makecell{LRS- \\ VQA}} & 
\multicolumn{3}{c}{\bfseries XLRS-Bench} & 
\multirow{2}[2]{*}{\bfseries Avg.} & 
\multirow{2}[2]{*}{\bfseries\makecell{\bm{$\Delta$} \\ Avg.}} \\ 
% 注意：上面最后一列使用了 \bm{\Delta} 来加粗希腊字母
\cmidrule(lr){7-9}
% 第二行表头 (XLRS-Bench 的子列)
& & & & & & 
\bfseries OSR & \bfseries CR & \bfseries Overall & & \\ 
\midrule
    Baseline  & \xmark & \xmark & \xmark & \xmark & 26.68 & 31.40 & 62.50 & 42.92 & 34.80 & -5.46 \\
    w/o GCC & \xmark & \cmark & \cmark & \cmark & 32.56 & 35.80 & 66.50 & 46.33 & 39.45 & -0.81 \\
    w/o MSES & \cmark & \xmark & \cmark & \cmark & 29.61 & 32.40 & 64.00 & 43.82 & 36.72 & -3.54 \\
    w/o SEM & \cmark & \cmark & \xmark & \cmark & 31.24 & 31.80 & 65.50 & 44.80 & 38.02 & -2.24 \\
    w/o TPEB & \cmark & \cmark & \cmark & \xmark & 28.86 & 31.00 & 63.00 & 43.29 & 36.08 & -4.18 \\
    \textbf{WeaveEarth (Ours)} & \cmark & \cmark & \cmark & \cmark & \textbf{33.38} & \textbf{36.40} & \textbf{67.50} & \textbf{47.14} & \textbf{40.26} & \textbf{-} \\
    \bottomrule
    \end{tabular}%
}
\caption{Ablation study results of WeaveEarth on LRS-VQA and XLRS-Bench benchmarks. The table reports the weighted average accuracy (\%) on LRS-VQA and XLRS-Bench, as well as the accuracy on specific sub-tasks Object Spatial Relationship (OSR) and Complex Reasoning (CR) in XLRS-Bench. "Avg." represents the average accuracy across both datasets, while the last column "$\Delta$ Avg." denotes the performance drop relative to the full WeaveEarth model.}
  \label{table:3}%
\end{table*}

\subsection{Ablation Study}
Table~\ref{table:3} presents the ablation results of WeaveEarth on LRS-VQA and XLRS-Bench. The results show that the performance gains of WeaveEarth do not come from any single module alone, but rather from the synergy between globally constrained evidence construction and structured evidence organization. In particular, the larger gains mainly arise from the subsequent stages of evidence compression, spatial grounding, and topology-preserving organization, rather than solely from the global-context constraint introduced at the candidate construction stage.

From the perspective of functional roles, the Global Context Constraint (GCC) mainly serves as a global prior during candidate evidence construction, corresponding to the second term, $\lambda \operatorname{sim}(p_i, g)$, in Eq.~(1), which constrains local regions to be consistent with the overall scene layout. Its effect is relatively modest, indicating that global context is beneficial for candidate proposal, but is not the primary bottleneck in performance. In contrast, removing the Minimal Support Evidence Set (MSES) leads to an average performance drop of 3.54 percentage points, showing that, in UHR scenarios, simply retaining more semantically relevant local regions does not naturally translate into better reasoning. Instead, redundant local regions dilute key clues, whereas a compact, low-redundancy, and spatially complementary support set is more beneficial for subsequent decision-making.

Removing Structured Evidence Metadata (SEM) causes an average drop of 2.24 percentage points, indicating that preserving local appearance alone is insufficient and that explicit bounding boxes, adjacency relations, and role attributes remain necessary for effective spatial grounding. Removing the Topology-Preserving Evidence Board (TPEB) results in the largest drop across the two datasets, reducing the average accuracy by 4.18 percentage points. This suggests that preserving relative topology and spatial adjacency among evidence regions is critical for effective reasoning.

When MSES or TPEB is removed, the performance on OSR and CR in XLRS-Bench drops substantially. This suggests that the difficulty of spatial reasoning lies not only in whether the model can find a relevant local target, but also in whether it can simultaneously access complementary evidence, explicit spatial attributes, and relative layout relationships. Here, SEM primarily improves patch-level spatial grounding, whereas TPEB mainly enhances set-level joint perception and spatial organization. Together with MSES, they constitute the key factors that distinguish WeaveEarth from simple patch retrieval.

\subsection{Plug-and-Play Transfer Across Frozen VLMs}

\begin{table}[t]
\small
\centering
% 使用 tabularx 环境，宽度设置为 \linewidth (填满单栏)
% X 列类型会自动计算宽度并平均分配剩余空间
\begin{tabularx}{\linewidth}{lcccc}
\toprule
\textbf{Method} & \textbf{LRS-VQA} & \textbf{MME} & \textbf{XLRS} & \textbf{Avg.} \\
\midrule
\rowcolor{row-pink1}\multicolumn{5}{c}{\textit{\textbf{Qwen3-VL-8B}}} \\
% Base 行保持不变
Baseline & \makebox[4.45em][c]{$28.16$} & \makebox[4.45em][c]{$42.11$} & \makebox[4.45em][c]{$42.92$} & \makebox[4.45em][c]{$37.73$} \\
% --- 修改部分开始 ---
\rowcolor{qwen7Highlight} 
\textbf{+WeaveEarth} & 
\makebox[1.8em][c]{$\textcolor{qwen7Text}{\textbf{33.38}}$}\,\makebox[0pt][l]{\textsuperscript{\textcolor{qwen7Text}{\scriptsize $\uparrow$5.22}}} & 
\makebox[1.8em][c]{$\textcolor{qwen7Text}{\textbf{47.38}}$}\,\makebox[0pt][l]{\textsuperscript{\textcolor{qwen7Text}{\scriptsize $\uparrow$5.27}}} & 
\makebox[1.8em][c]{$\textcolor{qwen7Text}{\textbf{47.14}}$}\,\makebox[0pt][l]{\textsuperscript{\textcolor{qwen7Text}{\scriptsize $\uparrow$4.22}}} & 
\makebox[1.8em][c]{$\textcolor{qwen7Text}{\textbf{42.63}}$}\,\makebox[0pt][l]{\textsuperscript{\textcolor{qwen7Text}{\scriptsize $\uparrow$4.90}}} \\
% Rel. 行
\rowcolor{qwen7Bg!40}
\textit{Relative} & 
\textcolor{qwen7Text}{\gain{33.38}{28.16}} & 
\textcolor{qwen7Text}{\gain{47.38}{42.11}} & 
\textcolor{qwen7Text}{\gain{47.14}{42.92}} & 
\textcolor{qwen7Text}{+13.0\%} \\
% --- 修改部分结束 ---
\midrule
\rowcolor{row-green}\multicolumn{5}{c}{\textit{\textbf{LLaVA-v1.6-7B}}} \\
Baseline & 23.55 & 23.39 & 24.00 & 23.65 \\
\rowcolor{qwenGreenHighlight} 
\textbf{+WeaveEarth} & 
\makebox[1.8em][c]{$\textcolor{qwenGreenText}{\textbf{27.60}}$}\,\makebox[0pt][l]{\textsuperscript{\textcolor{qwenGreenText}{\scriptsize $\uparrow$4.05}}} & 
\makebox[1.8em][c]{$\textcolor{qwenGreenText}{\textbf{28.04}}$}\,\makebox[0pt][l]{\textsuperscript{\textcolor{qwenGreenText}{\scriptsize $\uparrow$4.65}}} & 
\makebox[1.8em][c]{$\textcolor{qwenGreenText}{\textbf{28.87}}$}\,\makebox[0pt][l]{\textsuperscript{\textcolor{qwenGreenText}{\scriptsize $\uparrow$4.87}}} & 
\makebox[1.8em][c]{$\textcolor{qwenGreenText}{\textbf{28.17}}$}\,\makebox[0pt][l]{\textsuperscript{\textcolor{qwenGreenText}{\scriptsize $\uparrow$4.52}}} \\
\rowcolor{qwenGreenBg!40}
\textit{Relative} & 
\textcolor{qwenGreenText}{\gain{27.60}{23.55}} & 
\textcolor{qwenGreenText}{\gain{28.04}{23.39}} & 
\textcolor{qwenGreenText}{\gain{28.87}{24.00}} & 
\textcolor{qwenGreenText}{\gain{28.17}{23.65}} \\
\midrule
\rowcolor{row-blue1}\multicolumn{5}{c}{\textit{\textbf{IXC-2.5-7B}}} \\
Baseline & 28.27 & 36.12 & 30.00 & 31.46 \\
\rowcolor{qwen15Highlight} 
\textbf{+WeaveEarth} & 
\makebox[1.8em][c]{$\textcolor{qwen15Text}{\textbf{32.58}}$}\,\makebox[0pt][l]{\textsuperscript{\textcolor{qwen15Text}{\scriptsize $\uparrow$4.31}}} & 
\makebox[1.8em][c]{$\textcolor{qwen15Text}{\textbf{39.96}}$}\,\makebox[0pt][l]{\textsuperscript{\textcolor{qwen15Text}{\scriptsize $\uparrow$3.84}}} & 
\makebox[1.8em][c]{$\textcolor{qwen15Text}{\textbf{34.37}}$}\,\makebox[0pt][l]{\textsuperscript{\textcolor{qwen15Text}{\scriptsize $\uparrow$4.37}}} & 
\makebox[1.8em][c]{$\textcolor{qwen15Text}{\textbf{35.64}}$}\,\makebox[0pt][l]{\textsuperscript{\textcolor{qwen15Text}{\scriptsize $\uparrow$4.18}}} \\
\rowcolor{qwen15Bg!40}
\textit{Relative} & 
\textcolor{qwen15Text}{\gain{32.58}{28.27}} & 
\textcolor{qwen15Text}{\gain{39.96}{36.12}} & 
\textcolor{qwen15Text}{\gain{34.37}{30.00}} & 
\textcolor{qwen15Text}{\textbf{\gain{35.64}{31.46}}} \\
\bottomrule
\end{tabularx}
\caption{Plug-and-play transferability of WeaveEarth across multiple frozen VLM backbones. The evaluation metric is accuracy (\%). ``Avg.'' denotes the average accuracy across the three benchmarks. The superscript $\uparrow$ indicates the absolute improvement over the baseline, while the ``Relative'' row reports the relative improvement in percentage.}
\label{table:4}
\end{table}

To further verify that the performance gains of WeaveEarth do not rely on any single strong backbone, we instantiate it on two additional frozen backbones, LLaVA-v1.6-7B and IXC-2.5-7B. As shown in Table~\ref{table:4}, WeaveEarth consistently improves the performance of all three backbones across all benchmarks, with stable gains across backbones, yielding average improvements of +13.0\% for Qwen3-VL-8B, +19.1\% for LLaVA-v1.6-7B, and +13.3\% for IXC-2.5-7B, respectively. These results support the plug-and-play nature of WeaveEarth and suggest that its gains stem from the structured evidence construction and reasoning mechanism itself, rather than from any particular backbone.

\subsection{Efficiency and Budget Analysis}

\begin{figure}[t]
  \includegraphics[width=1\linewidth]{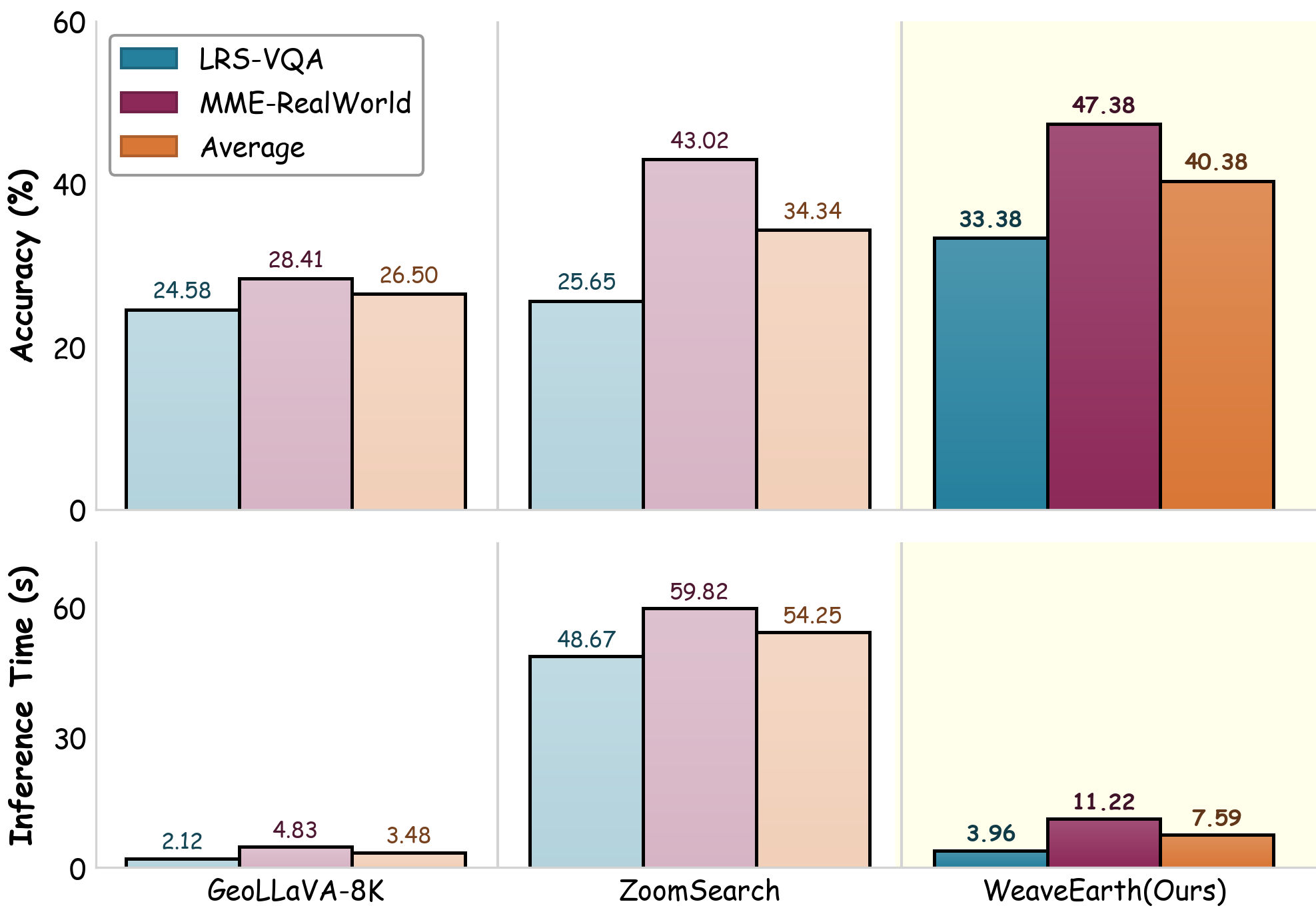}
  \caption{Accuracy-efficiency comparison of WeaveEarth with two representative UHR RS image understanding methods. GeoLLaVA-8K represents the passive perception paradigm, while ZoomSearch represents the active perception paradigm. The figure reports the accuracy (\%) on LRS-VQA, MME-RealWorld, and their average, together with per-sample inference time (s), in order to evaluate the trade-off between performance and inference cost across different methods.}
  \label{fig:3}
\end{figure}

\textbf{Accuracy-Efficiency Trade-off.} Figure~\ref{fig:3} compares WeaveEarth with two representative paradigms for UHR RS image understanding in terms of accuracy and inference time. Overall, WeaveEarth achieves a better balance between performance and inference cost.

Compared with GeoLLaVA-8K, a representative \emph{passive perception} method, WeaveEarth increases the average inference time on LRS-VQA and MME-RealWorld, but delivers a gain of 13.88 percentage points at the cost of only 4.11 seconds more per sample.

Compared with ZoomSearch, a representative \emph{active perception} method, WeaveEarth not only achieves a higher average accuracy (40.38\% vs. 34.34\%), but also reduces the average inference time from 54.25 s to 7.59 s. This result suggests that, rather than progressively discovering evidence through multi-round zooming, constructing key evidence under global context constraints first and then organizing it into a unified structured interface enables more stable UHR reasoning at a much lower cost.

WeaveEarth neither attempts to expand the model's visual access nor relies on repeated search. Instead, it compresses and organizes a small number of critical and complementary local evidence regions into an input interface that is well suited for frozen VLM reasoning. Therefore, Figure~\ref{fig:3} does not merely show that WeaveEarth is more efficient or more accurate; more importantly, it indicates that the key to UHR RS image understanding lies not in expanding visual access, but in improving the quality of evidence organization.

\begin{figure}[t]
  \includegraphics[width=1\linewidth]{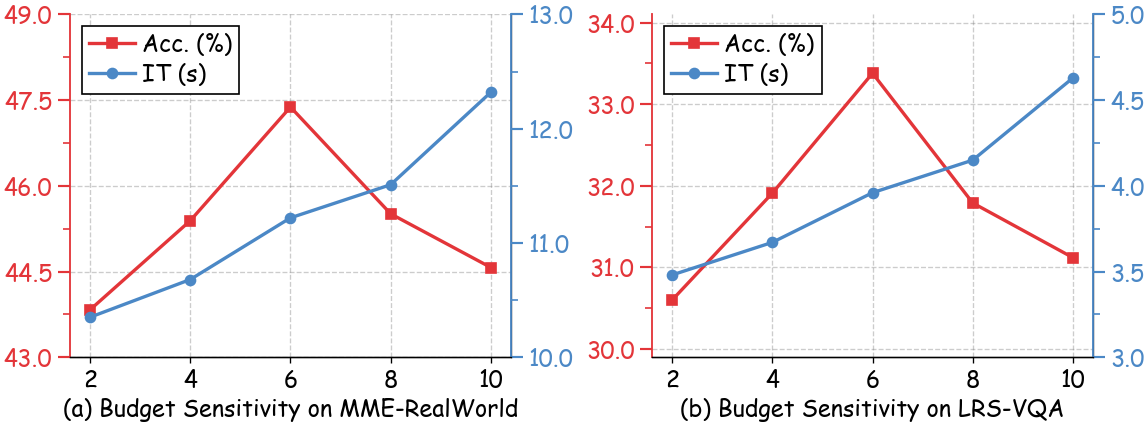}
  \caption{Performance and inference time under different evidence budgets on MME-RealWorld and LRS-VQA. The figure reports Accuracy (Acc., \%) and per-sample Inference Time (IT, s) as the support evidence budget varies from 2 to 10, in order to analyze the impact of evidence quantity on the performance and cost of WeaveEarth.}
  \label{fig:4}
\end{figure}

\noindent\textbf{Evidence Budget Sensitivity.} Figure~\ref{fig:4} further reports the effect of the evidence budget on performance and time cost. On both MME-RealWorld and LRS-VQA, accuracy consistently improves as the budget increases from 2 to 6, while inference time increases only slightly. However, when the budget is further increased to 8 and 10, accuracy begins to decline, whereas inference time continues to rise. In both experiments, the best performance is achieved at the default setting of budget = 6.

This result reveals a key phenomenon: for UHR RS image understanding, more evidence is not necessarily better; instead, there exists a minimal yet sufficient effective range. When the budget is too small, the support set fails to cover the critical clues required for reasoning. When the budget is too large, the additional regions introduce redundancy and distractions, weakening the model's ability to focus on the most informative evidence. Therefore, if the performance gain mainly came from expanding visual access, the accuracy should continue to improve as the evidence budget increases. However, the actual results exhibit a clear rise-then-fall trend, with the optimum achieved at the default budget of 6. This indicates that the gains of WeaveEarth do not come from simply enlarging the visual input, but from compressing, filtering, and structurally organizing evidence under limited budgets.

\subsection{Case Study and Discussion}

\begin{figure}[t]
  \includegraphics[width=1\linewidth]{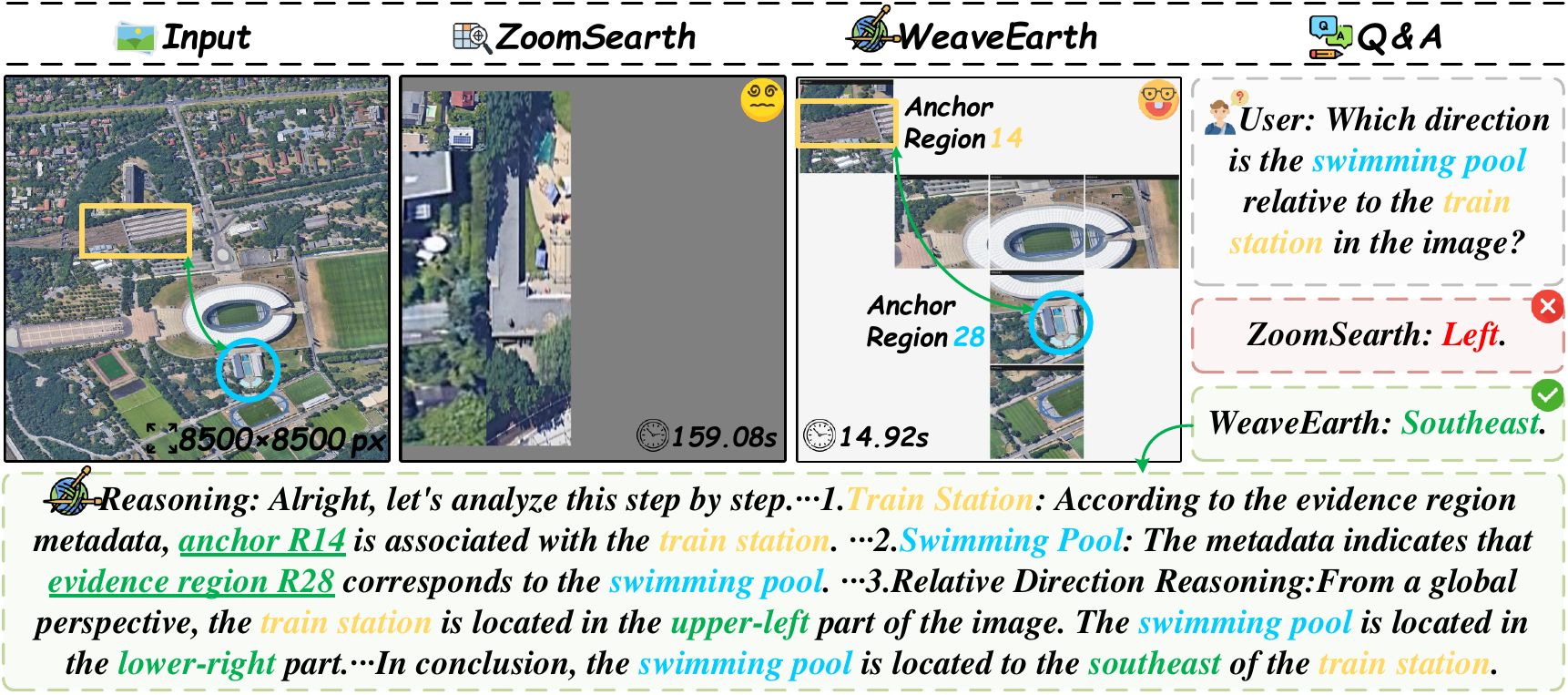}
  \caption{A case study on a UHR image. Due to mismatched and fragmented local evidence, ZoomSearch produces an incorrect answer. In contrast, WeaveEarth selects the correct key regions and answers correctly by organizing them into a globally grounded structured evidence interface, while also incurring much lower latency.}
  \label{fig:5}
\end{figure}

\noindent\textbf{Qualitative Case Study.} As shown in Figure~\ref{fig:5}, for the cross-region spatial relation question, ``What is the direction of the swimming pool relative to the train station?'', the failure of ZoomSearch arises from two aspects. First, it fails to stably localize the true key regions corresponding to the train station and the swimming pool. Second, even when the cropped local results are concatenated, this fragmented evidence still lacks reliable global orientation and relative layout information, which ultimately leads to an incorrect answer, with an inference time as high as 159.08 s.

In contrast, guided by the global thumbnail, WeaveEarth successfully identifies the anchor regions corresponding to the train station and the swimming pool, and answers the question correctly in a single round of reasoning through structured metadata and topology-preserving evidence organization, with an inference time of only 14.92 s. This case shows that the key to UHR RS reasoning lies not merely in discovering local evidence, but in whether critical evidence can be accurately selected and structurally organized under global context. Additional qualitative examples are provided in the appendix.

\noindent\textbf{Limitation and Future Work.} The model is still prone to errors on counting tasks. The main reason is that the objects to be counted in UHR RS images are often very small and densely distributed, making individual instances difficult to recognize reliably in the global thumbnail. Therefore, although WeaveEarth has already improved counting performance substantially over the corresponding baseline, its support for fine-grained instance counting remains limited. In future work, we plan to introduce multi-scale evidence modeling for counting tasks, so that the structured evidence interface can better cover scenes containing dense small objects.

\section{Conclusion}
In this paper, we propose WeaveEarth, a training-free framework for UHR RS image understanding via structured evidence construction and reasoning. Unlike whole-image perception or multi-round search, WeaveEarth reformulates the task as constructing key supporting evidence under global context constraints and performing joint reasoning through a structured evidence interface. Specifically, WeaveEarth uses Global-Aware Evidence Construction to build a Minimal Support Evidence Set, and Structured Evidence Reasoning to organize local evidence, spatial metadata, and relative topology into a unified input interface, thereby enhancing the global-local joint reasoning capability of frozen VLMs.

Extensive experiments demonstrate that WeaveEarth achieves consistent gains across multiple benchmarks and backbones, and further verify that its advantages mainly come from the compression, selection, and structured organization of critical evidence, rather than from expanding visual access. Therefore, for UHR RS image understanding, a more effective path lies not in accessing more, but in organizing better.

\bibliographystyle{ACM-Reference-Format}
\bibliography{WeaveEarth}

\end{document}